\documentclass{article}

\usepackage[nonatbib,preprint]{neurips_2023}

\usepackage{times}
\usepackage{epsfig}
\usepackage{graphicx}
\usepackage{amsmath}
\usepackage{amssymb}
\usepackage{hyperref}

\title{Conditional Generation from Unconditional Diffusion Models using Denoiser Representations}

\author{Alexandros Graikos \qquad Srikar Yellapragada \qquad Dimitris Samaras \\
\{agraikos, myellapragad, samaras\}@cs.stonybrook.edu \\
 Stony Brook University\\
 NY, USA
}

\newcommand{\vect}[1]{\mathbf{#1}}
\newcommand{\vectsymb}[1]{\boldsymbol{#1}}
\newcommand{\mat}[1]{\mathbf{#1}}
\newcommand{\normal}{\mathcal{N}}
\newcommand{\expect}{\mathbb{E}}
\newcommand{\norm}[1]{\lVert #1 \rVert}

\begin{document}

\maketitle

\begin{abstract}
Denoising diffusion models have gained popularity as a generative modeling technique for producing high-quality and diverse images. Applying these models to downstream tasks requires conditioning, which can take the form of text, class labels, or other forms of guidance. However, providing conditioning information to these models can be challenging, particularly when annotations are scarce or imprecise. In this paper, we propose adapting pre-trained unconditional diffusion models to new conditions using the learned internal representations of the denoiser network. We demonstrate the effectiveness of our approach on various conditional generation tasks, including attribute-conditioned generation and mask-conditioned generation. Additionally, we show that augmenting the Tiny ImageNet training set with synthetic images generated by our approach improves the classification accuracy of ResNet baselines by up to 8\%. Our approach provides a powerful and flexible way to adapt diffusion models to new conditions and generate high-quality augmented data for various conditional generation tasks.
\end{abstract}

\section{Introduction}

Denoising diffusion models have recently gained significant attention in the generative modeling literature due to their impressive results on various synthesis tasks, including image, audio, and molecule synthesis \cite{sohl2015deep, ho2020denoising, dhariwal2021diffusion, nichol2022glide, kong2021diffwave, hoogeboom2022equivariant}. Providing conditioning information can improve the sample quality of diffusion models, effectively guiding the model towards generating more diverse and representative samples \cite{ho2022cascaded, nichol2021improved, chen2022analog}.

While text-based conditioning has been widely used in image generation tasks, due to the wide availability of image-caption pairs, it is not always the best approach, as the scale at which the sampling guidance is applied can vary between tasks. For instance, sampling images with a dense per-pixel condition, such as a semantic mask, cannot be adequately substituted with a text condition, no matter how detailed. Additionally, it is often infeasible to provide finer annotations on the image or pixel level due to the large scale of the data used to train diffusion models. Thus, adapting pre-trained diffusion models to new conditions using the learned intermediate representations of the denoiser network is an attractive alternative.

In this paper, we propose a method to adapt pre-trained unconditional diffusion models to new conditions using the internal representations of the denoiser network. We show that the learned representations are inherently robust to noisy inputs, allowing us to provide guidance during sampling while utilizing a noisy estimate of the final image $\vect{x}_0$. Whereas previous methods relied on training large-scale guidance classifiers on the intermediate noisy steps, our method can efficiently learn from a small set of examples by exploiting the existing parameters of the denoiser network. We demonstrate the effectiveness of our approach on various conditional generation tasks, including attribute-conditioned generation and mask-conditioned generation.

When we are presented with more data but not enough to train a conditional diffusion model from scratch, we demonstrate that we can combine the learned unconditional model representations with the fine-tuning of the model. In particular, we focus on synthetic data augmentation, where we generate augmented data using a conditional diffusion model fine-tuned on Tiny ImageNet \cite{le2015tiny}. We use the internal representations of the unconditional U-Net \cite{ronneberger2015unet} denoiser network to train a rejection classifier. This classifier is used to reject low-quality samples generated by the fine-tuned conditional diffusion model, which helps to improve the overall quality of the generated images. We then use the augmented data to train a classification model and evaluate its performance on the Tiny ImageNet dataset. Our contributions are as follows: 
\begin{itemize}
    \item We demonstrate how the internal representations of an unconditional diffusion denoiser network can be used to adapt to new conditions with limited examples.
    \item We verify the effectiveness of our approach on various conditional generation tasks such as attribute-conditioned generation and mask-conditioned generation.
    \item We show how augmenting the Tiny ImageNet training set with synthetic images generated by our approach significantly improves the classification accuracy over ResNet \cite{he2015deep} baselines.
\end{itemize}

\section{Related Work}
\subsection{Conditional diffusion models}
Similar to other generative models, conditional diffusion models perform better than their unconditional counterparts, showing impressive results in text-conditioned generation \cite{nichol2022glide,rombach2022highresolution}. Apart from text-to-image models, in \cite{dhariwal2021diffusion}, the authors demonstrate how exploiting the denoiser network as a learned score function can be used with an auxiliary classifier trained on noisy samples to guide inference with class labels. In \cite{graikos2022diffusion}, the authors show that diffusion models can perform conditional generation by minimizing an energy function defined using the learned denoiser and any auxiliary constraint on the inferred sample.

\subsection{Conditioning of unconditional diffusion models}
Few-shot conditioning settings have utilized latent diffusion models in the past. In \cite{sinha2021d2c, preechakul2022diffusion}, the authors propose learning an expressive latent representation that can be utilized to condition diffusion models on a small set of labeled examples. The idea behind their approach is that the compressed latent representation can be efficiently used to adapt to new conditions. More recently, \cite{mou2023t2i,zhang2023adding} showcase faster and more efficient adaptation of diffusion models to new conditions, but both require a large number of data to train with.

\subsection{Diffusion as a pre-training task}
In \cite{baranchuk2022labelefficient}, the authors introduced the idea of utilizing the learned representations of a denoiser network for downstream tasks by training a segmentation network on top of the denoiser features with as few as 20 labels. Similarly, in \cite{brempong2022denoising} and \cite{tursynbek2023unsupervised}, the denoising task is exploited as a pre-training task for learning semantically meaningful features that can be easily adapted for performing downstream dense prediction tasks.

\section{Background}
A Gaussian diffusion process \cite{sohl2015deep,ho2020denoising} can be viewed as a latent variable model of the form
\begin{equation}
    p_{\theta}(\vect{x}_0) = \int p(\vect{x}_T)\prod_{t=1}^T p_{\theta}^{(t)} (\vect{x}_{t-1} \mid \vect{x}_t)
    d\vect{x}_{1}d\vect{x}_{2}\dots d\vect{x}_{T}
    \label{eq:lvm}
\end{equation}
where $\vect{x}_0$ is the data, $\vect{x}_{1},\dots,\vect{x}_{T-1}$ are intermediate latent variables that represent noisy versions of $\vect{x}_0$, and $\vect{x}_T$ is the terminal state with $\vect{x}_T \sim \normal (\vect{0},\mat{I})$. The forward process gradually adds Gaussian noise to the data and is defined by a noise schedule $\alpha_t$
\begin{equation}
    q(\vect{x}_t \mid \vect{x}_{t-1}) := N(\vect{x}_t; \sqrt{a_t}\vect{x}_{t-1}, (1-a_t)\mat{I}).
    \label{eq:fw_process}
\end{equation}
The reverse process is defined by learned Gaussian transitions and gradually denoises the data, starting from $\vect{x}_T := \normal(\vect{x}_T; \vect{0},\mat{I})$
\begin{equation}
    p_{\theta}(\vect{x}_{t-1} \mid \vect{x}_t) := N(\vect{x}_{t-1}; \vectsymb{\mu}_{\theta}(\vect{x}_t, t), \boldsymbol{\Sigma}_{\theta}(\vect{x}_t, t)).
    \label{eq:reverse_process}
\end{equation}
A property of the forward process is that it allows sampling of any intermediate $\vect{x}_t$ given $\vect{x}_0$, as
\begin{equation}
    q(\vect{x}_t \mid \vect{x}_0) = N(\vect{x}_t; \sqrt{\bar{a}_t}\vect{x}_0, (1-\bar{a}_t)\mat{I})
    \label{eq:xt_sampling}
\end{equation}
or equivalently
\begin{equation}
    \vect{x}_t = \sqrt{\bar{a}_t}\vect{x}_0 + \sqrt{1-\bar{a}_t}\vectsymb{\epsilon},\ \text{where}\ \vectsymb{\epsilon} \sim N(\vect{0},\mat{I}).
    \label{eq:xt_sampling_eq}
\end{equation}
with $\bar{a}_t = \prod_{i=1}^t a_i$. To approximate the reverse process, it is common to fix the variance $\boldsymbol{\Sigma}_{\theta} = \sigma_t \mat{I}$ and learn a function that predicts the noise added at each intermediate $\vect{x}_t$ by minimizing
\begin{equation}
    \expect_{\vect{x}_0, t} \left[ w(t) \norm{\epsilon - \epsilon_{\theta}(\sqrt{\bar{a}_t}\vect{x}_0 + \sqrt{1-\bar{a}_t}\vectsymb{\epsilon}, t)}^2 \right].
    \label{eq:ddpm_obj}
\end{equation}

The original DDPM \cite{ho2020denoising} formulation introduced sampling from a Gaussian diffusion model by sequentially applying the predicted noise
\begin{align}
    \vect{x}_{t-1} = \frac{1}{\sqrt{a_t}} 
    \left(\vect{x}_t  - \frac{1-\alpha_t}{\sqrt{1-\bar{\alpha}_t}} \vectsymb{\epsilon}_{\theta} (\vect{x}_t, t) \right)
    + \sigma_t \vect{z},\ \vect{z} \sim\normal(\vect{0},\mat{I})
    \label{eq:ddpm_sampling}
\end{align}
starting from a random $\vect{x}_T \sim N(\vect{0},\mat{I})$. Alternatively, in DDIM \cite{song2021denoising}, the authors proposed utilizing Eq. \ref{eq:xt_sampling_eq} to sample $\vect{x}_{t-1}$ from $\vect{x}_t$ by first estimating $\vect{x}_0$ and then using the known forward process to 'point-back' towards $\vect{x}_t$, as
\begin{align}
    \vect{x}_{t-1} = &\sqrt{\bar{\alpha}_{t-1}} 
    \underbrace{\left( \frac{\vect{x}_t - \sqrt{1-\bar{\alpha}_t} \vectsymb{\epsilon}_{\theta} (\vect{x}_t, t)}{\sqrt{\bar{\alpha}_t}}  \right)}_{\text{estimated } x_0} \nonumber
    + \underbrace{\sqrt{1 - \bar{\alpha}_{t-1} - \sigma_t^2}\vectsymb{\epsilon}_{\theta} (\vect{x}_t, t)}_{\text{direction pointing to } \vect{x}_t} \nonumber
    + \underbrace{\sigma_t \vect{z}}_{\text{random noise}}.
    \label{eq:ddim_sampling}
\end{align}

A denoiser $\vectsymb{\epsilon}_{\theta}(\vect{x}_t, t)$ can be interpreted as a learned score function of the noise-perturbed $\vect{x}_t$ \cite{dhariwal2021diffusion,song2021score}
\begin{equation}
    \nabla_{\vect{x}_t} \log p_{\theta} (\vect{x}_t) = -\frac{1}{\sqrt{1-\bar{\alpha}_t}} \vectsymb{\epsilon}_{\theta}(\vect{x}_t, t).
    \label{eq:score_func}
\end{equation}
For a conditioning variable $\vect{y}$, the score of the posterior $p(\vect{x}_t \mid \vect{y}) \propto p(\vect{x}_t)p(\vect{y} \mid \vect{x}_t)$ can be expressed as
\begin{align}
    \nabla_{\vect{x}_t} \log p(\vect{x}_t \mid \vect{y}) &= \nabla_{\vect{x}_t} \log p(\vect{x}_t) + \nabla_{\vect{x}_t} \log p(\vect{y} \mid \vect{x}_t) \nonumber
    \\&= -\frac{1}{\sqrt{1-\bar{\alpha}_t}}\vectsymb{\epsilon}_{\theta}(\vect{x}_t, t) + \nabla_{\vect{x}_t} \log p(\vect{y} \mid \vect{x}_t).
    \label{eq:posterior_score}
\end{align}
Thus by modifying the predicted noise $\vectsymb{\epsilon}_{\theta}$ at every step using the gradient of the log-likelihood w.r.t. $\vect{x}_t$ we can draw samples from the posterior distribution $p(\vect{x}_0 \mid \vect{y})$
\begin{equation}
    \hat{\vectsymb{\epsilon}}_{\theta}(\vect{x}_t, t) = \vectsymb{\epsilon}_{\theta} (\vect{x}_t, t) 
    - \lambda \sqrt{1-\bar{\alpha}_t} \nabla_{\vect{x}_t} \log p(\vect{y} \mid \vect{x}_t)
    \label{eq:modified_epsilon}
\end{equation}
where $\lambda$ is a tunable hyperparameter.

\section{Methodology}
In this section, we describe how we can utilize the learned intermediate representations of the denoiser network to conditionally generate and improve results from an unconditional diffusion model. 

\subsection{Learning the guidance signal using denoiser representations}
To provide conditional guidance during inference, we can compute the gradient of the log-likelihood of a condition $\vect{y}$ w.r.t. $\vect{x}_t$. Recent works employed a separate classifier $p(\vect{y} \mid \vect{x}_t)$ trained on \textbf{noisy} samples $\vect{x}_t$ to provide semantic guidance, in the form of class labels. This work argues that retraining the guidance network from scratch is unnecessary. Instead, we propose using the intermediate denoiser representations, with the estimate of $\vect{x}_0$ at every step, to guide sampling. 

Recall that from Eq. \ref{eq:xt_sampling_eq} we can compute an estimate of the final sample $\vect{x}_0$ from
\begin{equation}
    \hat{\vect{x}}_0 = \frac{\vect{x}_t - \sqrt{1-\bar{\alpha}_t} \vectsymb{\epsilon}_{\theta} (\vect{x}_t, t)}{\sqrt{\bar{\alpha}_t}}.
    \label{eq:x0_estimate}
\end{equation}
Using this point estimate $p(\vect{x}_0 \mid \vect{x}_t) = \delta (\vect{x}_0 - \hat{\vect{x}}_0)$ we can rewrite the likelihood as
\begin{align}
    p(\vect{y} \mid \vect{x}_t) &= \sum_{\vect{x_0}} p(\vect{y} \mid \vect{x}_t,\vect{x}_0) p(\vect{x}_0 \mid \vect{x}_t)
    \\&= \sum_{\vect{x_0}} p(\vect{y} \mid \vect{x}_0) p(\vect{x}_0 \mid \vect{x}_t) \tag*{(cond. ind.)}
    \\&= p(\vect{y} \mid \hat{\vect{x}}_0) \tag*{(point estimate)}
    \label{eq:x0_likelihood}
\end{align}
and plug-in Eq. \ref{eq:modified_epsilon} to get
\begin{equation}
    \hat{\vectsymb{\epsilon}}_{\theta}(\vect{x}_t, t) = \vectsymb{\epsilon}_{\theta} (\vect{x}_t, t) 
    - \sqrt{1-\bar{\alpha}_t} \nabla_{\vect{x}_t} \log p(\vect{y} \mid \vect{x}_0(\vect{x}_t)).
    \label{eq:modified_epsilon_x0}
\end{equation}

By expressing the likelihood in terms of an estimate of $\vect{x}_0$, we allow our guidance network to be trained only on 'clean' samples, thus reducing the overall complexity of the post-hoc adaptation process. However, this assumes that the estimates $\hat{\vect{x}}_0$ perfectly follow the marginal distribution $p_\theta(\vect{x}_0)$, which is not true, notably for the first few estimates in the denoising process. In practice, the guidance network $p(\vect{y} \mid \hat{\vect{x}}_0)$ will initially steer the sample $\vect{x}_t$ towards the wrong directions, which in most cases cannot be mitigated by the following denoising updates.

By using the unconditional denoiser network as a feature extractor, we can both provide guidance that is robust to the initial inaccurate estimates of $\vect{x}_0$ and quickly learn the guidance directions from a small set of labeled samples. The denoiser network is trained with varying noise levels in its inputs and has learned to extract features of different scales at different timesteps. Additionally, the intermediate features of the denoiser U-Net are shown to contain important information for downstream tasks \cite{baranchuk2022labelefficient}. Therefore, we argue that apart from robustness, using the denoiser features enables us to learn the conditioning models with a small set of examples. 

\begin{figure*}[t]
\centering
\includegraphics[width=0.95\linewidth]{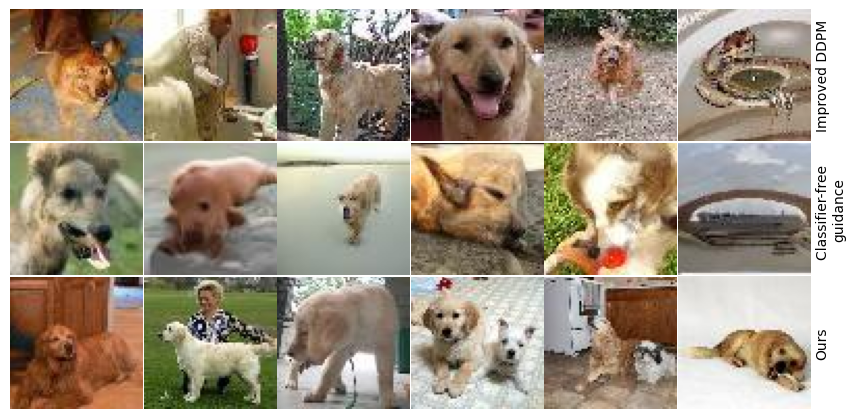}
   \caption{Synthetic images from the fine-tuned class conditional sampling (first row) vs Classifier-free guided sampling (second row) vs. our model with rejection (third row). Samples are from class 9: Golden Retriever on the Tiny-Imagenet. Our method generated more realistic and diverse images compared to others.}
\label{fig:tiny_images}
\end{figure*}

\subsection{Combining adaptation with denoiser representations}
When presented with more data, we can combine model adaptation methods with the unconditional denoiser features to generate higher-quality and diverse images. We begin by fine-tuning the unconditional diffusion model on the labeled examples we are given by naively adding the conditioning connections to the pre-trained model. It is important to note that here we assume that the image-label pairs that are provided do not cover the entirety of the dataset, which would make the task trivial.

Even with more image-condition pairs, the conditional model fails to generate realistic images, as demonstrated in Fig.~\ref{fig:tiny_images}. We use the unconditional denoiser representations to train a classifier of the labels of the given samples and improve the results by applying rejection sampling. The classifier model decides whether to accept a generated sample using the predicted probability value.

We find this method preferable to just providing guidance during sampling since tuning the weight of the guidance becomes increasingly difficult, with an inherent trade-off between quality and diversity that becomes infeasible to hand-tune. We leave the problem of controlling the guidance during sampling to future work.

\section{Experiments}
\subsection{Few-shot guidance for face attributes}
\label{sec:attr_experiment}
We first demonstrate how we can generate conditional samples with a limited number of image-attribute pairs. We utilize an unconditional diffusion model trained on the CelebA-64 \cite{liu2015faceattributes} dataset. Similarly to \cite{sinha2021d2c}, we train an attribute classifier by giving 50 positive and 50 negative examples of a single attribute, such as blonde or male. 

To train the classifier, we extract the bottleneck and second upsampling block features from the denoiser U-Net at $T=700$. The classifier network is a simple 3-layer convolutional network trained for 100 steps. During sampling, we apply classifier guidance with a weight of $\lambda=1$. The results for the conditional generation of images with the attributes \textit{male}, \textit{female}, \textit{blond}, and \textit{non-blond} are shown in Table \ref{tab:attr_experiment}. 

In order to compare with D2C \cite{sinha2021d2c} and DiffAE \cite{preechakul2022diffusion}, we train a classifier using the latent representation of each and generate samples by unconditionally sampling from the model and applying rejection sampling. When comparing with DiffAE and D2C, we show that we can provide meaningful guidance and achieve comparable FID scores without requiring to compress the entire image into a latent representation and without rejection, which significantly increases the computation needed. For DDIM-I, we use our approach for providing guidance with an estimate of $\hat{\vect{x}}_0$, but now using a separately trained classifier network trained only on ``clean'' images. The subpar performance validates the assumption that the intermediate denoiser representations are more robust to the initial inaccurate estimates of the final image. All FID scores are computed using the images of the CelebA-64 test set with the target attribute.

\begin{table}[t]
\begin{center}
\begin{tabular}{|l|c|c|c|c|c|c|}
    \hline
    Class & Ours & DiffAE \cite{preechakul2022diffusion} & D2C \cite{sinha2021d2c} & DDIM-I & NVAE \cite{vahdat2020nvae} \\
    \hline\hline
    Male & 15.34 & 11.52 & 13.44 & 29.03 & 41.07 \\
    Female & 9.94 & 7.29 & 9.51 & 15.17 & 16.57 \\
    \hline
    Blond & 13.07 & 16.10 & 17.61 & 29.09 & 31.24 \\
    Non-Blond & 10.97 & 8.48 & 8.94 & 19.76 & 16.73 \\
    \hline
\end{tabular}
\end{center}
\caption{FID scores for attribute-conditioned generation on the CelebA-64 dataset.}
\label{tab:attr_experiment}
\end{table}

\subsection{Few-shot guidance for semantic segmentations}
We also demonstrate how we can generate conditional samples with a small set of paired images and segmentation masks. We utilize the dataset of \cite{baranchuk2022labelefficient} and an unconditional diffusion model trained on FFHQ-256 \cite{karras2019style}. In this setting, we train on 20 images and their corresponding segmentation masks to generate realistic-looking, semantically accurate images. 

We again train a per-pixel classifier network using intermediate U-Net representations extracted at $T=500$ from decoder upsampling blocks. The classifier is a simple per-pixel multi-layer perceptron trained for 50 steps. During sampling, we apply classifier guidance of $\lambda=2\times10^{-4}$, which is significantly lower than for the single attribute case of \ref{sec:attr_experiment}, which we attribute to the finer per-pixel adjustments that we now have to make.

\begin{figure}[t]
\begin{center}
    \begin{tabular}{@{}c@{}c@{}c@{}c@{}c@{}c@{}}
    {\small Mask} & {\small Image} & {\small DDIM-I} & {\small DiffAE} \cite{preechakul2022diffusion}
    & {\small Ours} & {\small Ours - 100 steps} \\
    \includegraphics[width=0.16\linewidth]{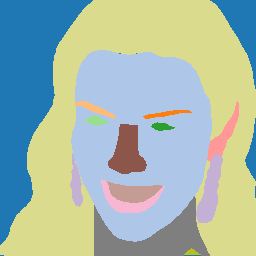} &
    \includegraphics[width=0.16\linewidth]{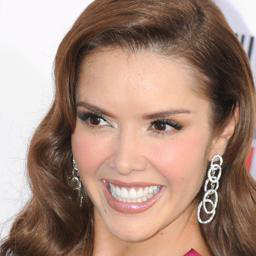} &
    \includegraphics[width=0.16\linewidth]{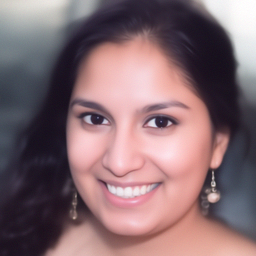} &
    \includegraphics[width=0.16\linewidth]{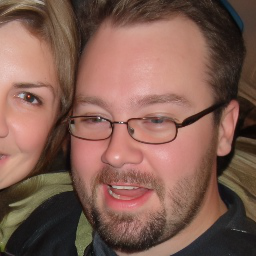} &
    \includegraphics[width=0.16\linewidth]{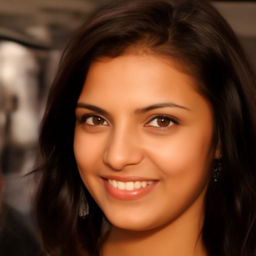} &
    \includegraphics[width=0.16\linewidth]{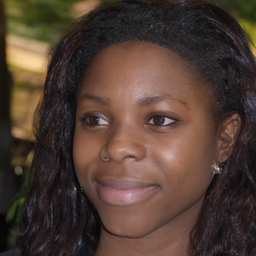} \\
    \end{tabular}
\end{center}
   \caption{Example of segmentation-conditioned generated images for each method.}
\label{fig:seg_experiment}
\end{figure}

In Table \ref{tab:seg_experiment}, we provide the results of generating images conditioned on segmentation masks taken from CelebA-Mask \cite{lee2020maskgan}. We compute the mIoU over all classes between the generated images and input segmentations using a separately trained Oracle segmentation model on a test set of 500 CelebA-Mask images. We also compute the FID scores between the two. When comparing to the naive approach of using a separately trained segmentation network to provide guidance using the estimates of $\hat{\vect{x}}_0$, we see that although we get slightly correct semantic results, the quality is reduced. On the contrary, when using the DiffAE latent to train a segmentation network and sample with rejection sampling, we can improve the quality at the loss of semantic correctness and speed. The latent representation compresses the image information in that case and thus cannot accurately project the segmentation onto the entire image. Our method provides more diverse and high-quality samples without sacrificing semantic correctness. We also note that we run our method with only ten denoising steps to make the comparison to the faster DiffAE fair. If we can perform more steps, the results can improve, as shown in Fig.~\ref{fig:seg_experiment}.

\begin{table}[t]
\begin{center}
\begin{tabular}{|l|c|c|}
    \hline
    Model & mIoU $\uparrow$ & FID $\downarrow$ \\
    \hline\hline
    DDIM-I & 0.28 & 106.90 \\
    DiffAE \cite{preechakul2022diffusion} & 0.23 & 87.68 \\
    \textbf{Ours} & \textbf{0.31} & \textbf{74.74} \\
    \hline
\end{tabular}
\end{center}
\caption{Mean Intersection over Union and FID on the 500-image CelebA-Mask test set for a segmentation-conditioned generation.}
\label{tab:seg_experiment}
\end{table}

\subsection{Synthetic data augmentation}

\subsubsection{Setup}

To evaluate the effectiveness of our proposed approach, we conduct experiments on Tiny Imagenet \cite{tinyimagenet}. It has 200 object categories and 100,000 images, with 500 images per category for training and 10,000 images for validation. The images are of size 64x64 pixels. 

We start with an unconditional DDPM trained on Imagenet images with a cosine noise schedule and fine-tune it as a class-conditioned diffusion model by adding a classifier embedding as proposed in \cite{nichol2021improved}. The classifier embedding layer weights are randomly initialized, and we train this model for 150,000 steps with a batch size of 128 and a learning rate of 0.0001. 

Since the Tiny Imagenet dataset is much smaller than Imagenet, the resulting conditional model fails to generate realistic images, as seen in Figure \ref{fig:tiny_images}. As a solution, we also train a rejection classifier to improve the quality and diversity of generated images. We extract Tiny-Imagenet features from the unconditional DDPM's U-Net bottleneck layer. Two feature vectors extracted at timestep $T=300$ and $T=800$ are  concatenated to obtain a $1024\times4\times4$ embedding. The rejection classifier is a 4-layer CNN with 1.2M parameters built on top of the U-Net features.  We set a threshold of 0.2 and discard any generated image for which the classifier predicts a probability lower than the threshold.

\begin{figure*}[t]
\centering
\includegraphics[width=0.95\linewidth]{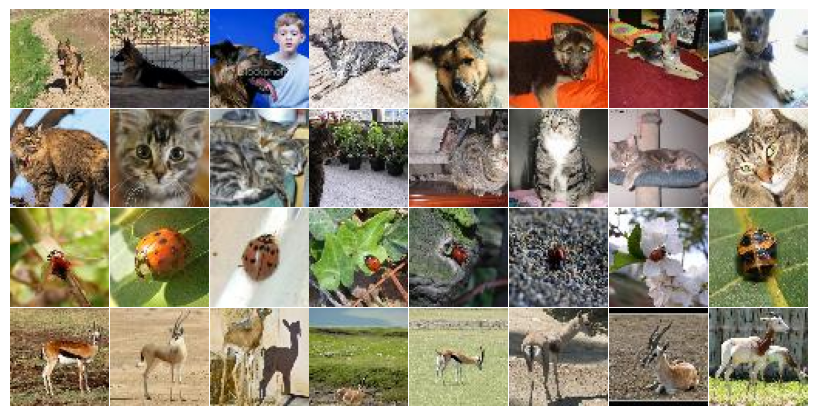}
   \caption{Tiny-Imagenet samples generated by our method. The classes are 28: German Shepherd, 30: Tabby Cat, 36: Ladybug, and 52: Gazelle. Our model can generate diverse images across different classes, proving the effectiveness of rejection sampling.}
\label{fig:tiny_images_ours}
\end{figure*}

We use classifier-free guidance sampling \cite{ho2021classifier} to generate images from our class conditional model, taking a linear combination of unconditional and conditional estimates. We use a low guidance weight of 0.01 to encourage higher diversity and generate additional synthetic images for the classes in Tiny-Imagenet to construct training datasets with $100,000-400,000$ synthetic images. Figure \ref{fig:tiny_images_ours} shows sample images generated by our approach.

\subsubsection{Classification Accuracy Score}
\label{expt-cas}
Classification Accuracy Score (CAS) \cite{ravuri2019classification} can be a better proxy than Inception and FID for measuring the quality of synthetic datasets. CAS was originally designed for Imagenet; we adopted it for Tiny-Imagenet by computing the Top-1 accuracy on the Tiny-Imagenet validation set using a classifier trained on the synthetic data. 

Table \ref{tab:cas} reports the CAS for our synthetic data. In all cases, we train a ResNet-18 model for 40 epochs with a batch size of 256 using the Adam optimizer \cite{kingma2014adam} on the different data sources. The learning rate was 0.001 for 20 epochs and reduced to 1/0th at epochs 20 and 30.  We generate synthetic data using the fine-tuned class conditional DDPM (Improved DDPM), combining the unconditional and conditional models (Classifier-free guidance), and using our method. The classifier trained on our synthetic data achieves a validation accuracy of \textbf{45.09 \%}, significantly more than the other baselines, showcasing our capability to generate more diverse and higher-quality data. 

\begin{table}[t]
\centering
\begin{tabular}{|l|c|c|}
\hline
Data type & Data Source                         & Top-1 accuracy (\%) \\ \hline \hline
Real      & Tiny-Imagenet                 & 52.24               \\ \hline
Synthetic & Improved DDPM \cite{nichol2021improved}                 &  35.37              \\
Synthetic & Classifier-free guidance \cite{ho2022classifier}       &  36.12              \\
Synthetic & \textbf{Ours} & \textbf{45.09}               \\ \hline
\end{tabular}
\caption{Classification Accuracy scores of a Resnet-18 on synthetic data. Results indicate rejection is critical in making our synthetic data effective.}
\label{tab:cas}
\end{table}

\subsubsection{Combining real and synthetic data}
\label{sub-training}
Having verified the quality of our synthetic data, we also try improving the baseline classification accuracy on Tiny-Imagenet by augmenting the real data with increasing amounts of diffusion-generated synthetic data. We follow the same training procedure as \ref{expt-cas} , using three network architectures: ResNet-18, Wide-ResNet-50 \cite{zagoruyko2016wide}, and ResNeXt-50 \cite{xie2017aggregated}. Results can be found in Table \ref{tab:tiny-imagenet}. 

We observe that augmenting the training set with our synthetic images improves the accuracy of ResNet baselines by up to \textbf{8\%}. ResNeXt-50 performs the best, improving accuracy from 53.98\% with real data only to \textbf{63.15\%} with the additional augmented data.  We see that performance continues to improve as we increase the amount of synthetic data, with the expected diminishing returns. The improvement is also consistent across all three network architectures, which validates the effectiveness of our data generation method.

\subsubsection{Comparison with image level augmentations}

To evaluate the effectiveness of our synthetic data compared to image-level augmentation methods, we conducted experiments on Tiny Imagenet  using Mixup \cite{zhang2017mixup} and Cutmix \cite{yun2019cutmix} techniques. We applied Mixup and Cutmix with a probability of 0.5 on all images in the training set (both real and synthetic), following the same training procedure from \ref{sub-training}. The results of these experiments are summarized in Table \ref{tab:tiny-imagenet}.

Our observations reveal that Mixup and Cutmix techniques further enhance the accuracy of ResNet baselines. This improvement is consistent across all the network architectures and different amounts of synthetic data. These findings demonstrate that our augmentation approach complements image-level augmentations and can be effectively combined to achieve even better performance.

\begin{table}[ht]
\centering
\begin{tabular}{|c|c|c|c|c|c|}
\hline
   & Mixup + &  & Real + & Real + & Real + \\ 
Architecture & Cutmix & Real only & 1x Generated & 2x Generated & 3x Generated \\ 
\hline\hline
Resnet-18  &  No  & 52.24     & 56.13                                                          & 58.13                                                         & \textbf{59.37}                                                          \\
  &  \textbf{Yes}  & 52.9     & 58.9                                                          & 62.01                                                         & \textbf{62.75}                                                          \\\hline

Wide-resnet-50 & No & 53.27     & 58.57                                                          & 61.71                                                          & \textbf{62.82}                                                          \\
 & \textbf{Yes} & 56.56     & 62.71                                                          & 66.42                                                          & \textbf{66.82}                                                          \\\hline

ResNeXt-50 &   No & 53.98     & 59.33                                                          & 62.27                                                          & \textbf{63.15}                                                          \\
 &  \textbf{Yes}  & 57.98     & 64.4                                                          & 66.85                                                          & \textbf{67.05}                                                          \\\hline
\end{tabular}
\caption{Comparison of Top-1 Accuracy (\%) with our synthetic data. When ``Mixup+Cutmix" is indicated as``Yes," it signifies that Mixup and Cutmix augmentation techniques were applied to the entire training set with a probability of 0.5. We can see that our augmentation approach complements image-level augmentations.}
\label{tab:tiny-imagenet}
\end{table}

\subsubsection{Diminishing returns}

\begin{figure}
    \centering
    \includegraphics[width=\linewidth]{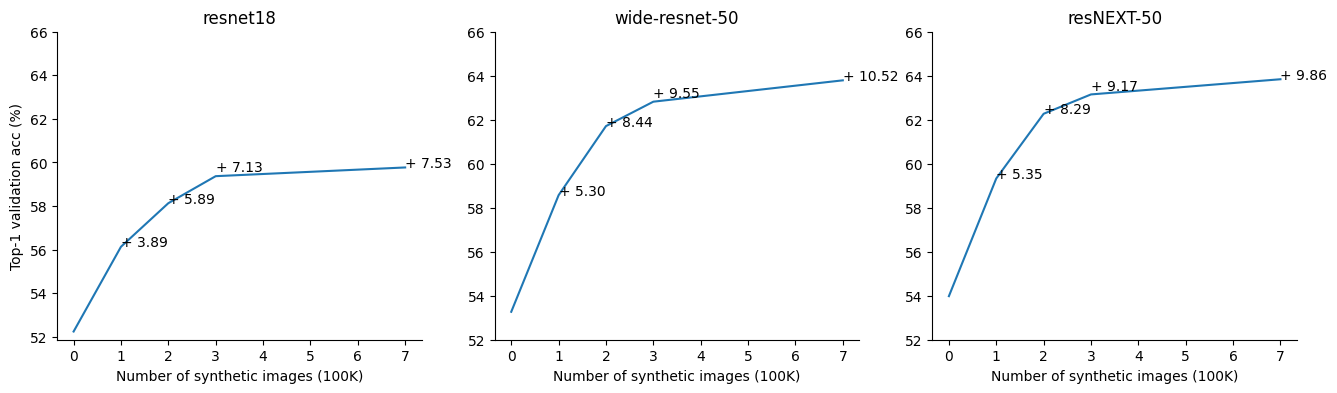}
    \caption{Improvement in Top-1 Accuracy (\%) with synthetic data. The X-axis represents the number of synthetic images used, scaled by a factor of 100k, with 0 indicating the utilization of only real images. As anticipated, the graph demonstrates diminishing returns as synthetic data increases.}
    \label{fig:my_label}
\end{figure}

When progressing from 3x generated data (300k synthetic images) to 7x (700k images), we observed diminishing returns in the improvement of validation accuracy, as depicted in Figure \ref{fig:my_label}. Initially, incorporating more synthetic images led to a notable increase in validation accuracy, indicating the benefits of augmented data. However, as we further scaled up, the extent of improvement diminished, suggesting a saturation point where additional synthetic data yielded diminishing gains, possibly indicating a limit to the model's ability to extract valuable information from the augmented samples.

\section{Conclusion}
In this paper, we proposed a method for adapting pre-trained unconditional diffusion models to new conditions using the internal learned representations of the denoiser network. The effectiveness of the proposed approach is demonstrated by providing guidance for attribute-conditioned generation and mask-conditioned generation, as well as filtering samples for synthetic data augmentation. We showed that augmenting the Tiny ImageNet training set with synthetic images generated by our approach significantly improves the classification accuracy over ResNet baselines. This result highlights the potential of our approach to improving the performance on datasets with limited labels by using a generative diffusion model as a pre-training task or feature extractor.

\bibliographystyle{abbrv}
\bibliography{submission}

\end{document}